\documentclass[10pt,twocolumn]{article}

\usepackage{cvpr}
\usepackage{times}
\usepackage{epsfig}
\usepackage{graphicx}
\usepackage{amsmath}
\usepackage{amssymb}
\graphicspath{ {./images/} }

\usepackage[a4paper, margin=2cm]{geometry}

\usepackage[breaklinks=true,bookmarks=false]{hyperref}

\cvprfinalcopy 

\def\cvprPaperID{****} 

\begin{document}

\title{COVID-19 Tweets Analysis through Transformer Language Models}
\author{Abdul Hameed Azeemi\\
Information Technology University\\
{\tt\small msds19003@itu.edu.pk}

\and
Adeel Waheed\\
Information Technology University\\
{\tt\small msds19030@itu.edu.pk}

}

\maketitle

\section{Abstract}
Understanding the public sentiment and perception in a healthcare crisis is essential for developing appropriate crisis management techniques. While some studies have used Twitter data for predictive modelling during COVID-19, fine-grained sentiment analysis of the opinion of people on social media during this pandemic has not yet been done. In this study, we perform an in-depth, fine-grained sentiment analysis of tweets in COVID-19. For this purpose, we perform supervised training of four transformer language models on the downstream task of multi-label classification of tweets into seven tone classes: [confident, anger, fear, joy, sadness, analytical, tentative]. We achieve a LRAP (Label Ranking Average Precision) score of 0.9267 through RoBERTa. This trained transformer model is able to correctly predict, with high accuracy, the tone of a tweet. We then leverage this model for predicting tones for 200,000 tweets on COVID-19. We then perform a country-wise analysis of the tone of tweets, and extract useful indicators of the psychological condition about the people in this pandemic.

\section{Introduction}
COVID-19 has affected more than 11 million people all around the world as of 28 May 2020. The overall public sentiment on Twitter during this global pandemic has largely echoed the real world events and the situation prevailing in that region. During COVID-19, people have extensively used social media platforms, particularly Twitter, for conveying medical information, conveying the stats, expressing emotions and alerting other people of the impending danger. 

In the past, global health events have shown that social media surveillance systems can be successfully utilized to extract the public sentiment and perception instantaneously \cite{ji2015twitter, ji2013monitoring, signorini2011use}. In this scenario, the biggest advantage of a social media platform such as Twitter is its global, instantaneous coverage, which makes it highly suitable for use in real-time, adaptive alert systems. 

Despite the emergence of a number of works on the social media analysis of COVID-19 tweets, fine-grained sentiment analysis still remains an unexplored area in this regard. Particularly, there is a need of a system that can reliably extract tone of the ever-growing number of COVID-19 tweets, and then use it to make reliable judgements about the public sentiment.

In this project, we tackle the problem of multi-label classification of tweets into multiple tone classes (angry, sad, analytical etc.). We train a transformer based language model on this task (RoBERTa) and use it for predicting the tone of a large number of tweets.

Github: \url{https://github.com/ahazeemi/MSDS19003_Project_DLSpring2020}

\section{Related Work}
A wealth of recent studies have utilized the tweets during this pandemic for extracting useful information and presenting insights into the public health. Particularly, sentiment analysis has been utilized in the analysis of lockdown life \cite{thelwall2020retweeting}, topic modelling has been used for the analysis of the response of politicians \cite{sha2020dynamic} and situation forecasting has been leveraged for surfacing the techniques of crisis management \cite{gharavi2020early} during COVID-19. 

A recent, useful study uses causal inference approach through bayesian networks to discover and quantify causal relationships between pandemic characteristics (e.g. number of infections and deaths) and Twitter activity as well as public sentiment \cite{gencoglu2020causal}. The authors use DistilBERT for sentiment analysis of the tweets. The model labels each tweet with a sentiment of POSTIVE or NEGATIVE. This sentiment is then related to the country-wide statistics for 12 countries: Italy, Spain, Germany,
France, Switzerland, United Kingdom, Netherlands, Norway, Austria, Belgium, Sweden, and
Denmark. 

This work is closely related to our study as it uses a transformer-based model (BERT) for analyzing COVID-19 tweets and then relating it to the country-wide statistics. However, the type of sentiment analysis in this paper produces a boolean output i.e. POSITIVE or NEGATIVE. On the other hand, our study relates to labelling of COVID-19 tweets with some of the seven tone classes: [confident, anger, fear, joy, sadness, analytical, tentative] through transformer based language models. Hence, our problem is essentially multi-label tweet classification through transformer language models.

Since we will be using transformer-based language models, we have selected four popular transformer LM models: BERT, RoBERTa, XLNet and ELECTRA which achieve high accuracy on natural language understanding tasks.
\\
\textbf{BERT} \cite{devlin2018bert}:
BERT is based on transformer architecture. It is designed to pre-train bidirectional representations from unlabeled text by jointly conditioning on both left and right context. As a result, the pre-trained BERT model can be fine-tuned with just one additional output layer to create state-of-the-art models for a wide range of NLP tasks. BERT is pre-trained on two NLP tasks. 
1.	Masked Language Models
2.	Next Sentence Prediction
\\
\textbf{ELECTRA} \cite{clark2020electra}:
ELECTRA (Efficiently Learning an Encoder that Classifies Token Replacements Accurately) is a new pre-training approach which aims to match or exceed the downstream performance of an MLM (Masked Language Model) pre-trained model while using significantly less compute resources for the pre-training stage. The pre-training task in ELECTRA is based on detecting replaced tokens in the input sequence. This setup requires two Transformer models, a generator and a discriminator.
\\
\textbf{ROBERTA} \cite{liu2019roberta}:
RoBERTa is an optimized BERT Pretraining Approach. RoBERTa performs better than BERT by applying the following adjustments:
\begin{enumerate}
    \item RoBERTa uses BookCorpus (16G), CC-NEWS (76G), OpenWebText (38G) and Stories (31G) data while BERT only uses BookCorpus as training data only. 
    \item BERT masks training data once for MLM objective while RoBERTa duplicates training data 10 times and masks this data differently.
\end{enumerate}

\textbf{XLNet} \cite{yang2019xlnet}:
XLNET is a generalized autoregressive (AR) model where next token is dependent on all previous tokens. XLNET is generalized because it captures bi-directional context by means of a mechanism called permutation language modeling. AR language model is a kind of model that using the context word to predict the next word. But here the context word is constrained to two directions, either forward or backward. BERT outperforms previous LMs but XLNET outperforms BERT. It uses the [MASK] in the pretraining but this kind of symbols are absent from real data at fine-tuning time resulting in a pretrain-finetune discrepancy. XLNET proposes a new a way to avoid the disadvantages brought by the [MASK] method in BERT. In pre-train phase, XLNET proposed a new objective called Permutation Language Modeling. This objective learns contextual text representation using permutation of input.

\section{Dataset}
The dataset consists of 658,967 COVID-19 tweets of seven days (from 25-03-2020 to 31-03-2020). The tweets in the dataset were extracted using these hashtags: ["isolation", "social distance", "socialdistancing", "socialdistancing", "confined", "stayathome",  "covid", "stayathome", "sath", "corona", "covid-19", "quarantine", "isolation", "untiltomorow", "homeoffice"]. This dataset contains the following columns: user-id, tweet, tweet-id, followers, location. The data cleanup and tone extraction methodology is as under:
\begin{enumerate}
    \item Only the tweets with retweet count $>$ 1 were kept. This removes most of the bot-generated tweets and results in a high-quality dataset. The count of the resulting tweets is 20,200.
    \item The tweets were grouped by the date they were posted, and randomly 2000 tweets were extracted from each day. The resulting tweets were 12,461 in number.
    \item The tone for each of these tweets was extracted using the IBM Watson Tone Analyzer API. The API tags a piece of text with one of these seven tone classes: [confident, anger, fear, joy, sadness, analytical, tentative]. A tone is only assigned to a text if it has been predicted with high probability ($>$ 0.5).
\end{enumerate}

\begin{table}
\begin{center}
\begin{tabular}{|l|c|}
\hline
Tone & Number of Tweets\\
\hline\hline
Anger & 268 \\
Analytical & 4364 \\
Tentative & 3136 \\
Confident & 2373 \\
Sadness & 1405 \\
Fear & 287\\
Joy & 4274\\
\hline
\end{tabular}
\end{center}
\caption{Number of tweets belonging to each tone. The total number of tweets is 12,461 }
\end{table}

\begin{figure}[h!]
  \includegraphics[width=8cm]{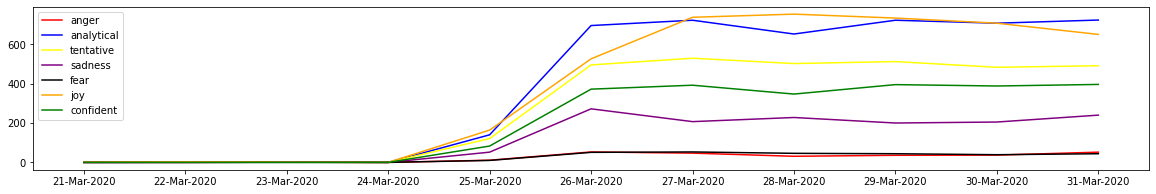}
  \caption{The number of tweets of each tone per day} 
\end{figure}

\section{Methodology}
We leverage transfer learning for training a model that can accurately predict the tone of any given text. There are two steps in the application of transfer learning in NLP:
\begin{enumerate}
    \item Unsupervised training on large amounts of text (e.g. wikipedia, books etc.) The output of this step is a trained model (e.g. BERT) which has captured the patterns of a language like english.
    \item Supervised finetuning on a downstream task with a labelled dataset e.g. text classification, sentiment analysis etc. 
\end{enumerate}

In our case, we selected four transformer based models that had already been trained through step 1: RoBERTa, Electra, XLNet, BERT. We then performed supervised finetuning of these transformer models on the downstream task of tone classification into seven classes. This part essentially adapts the transformer parameters to the supervised target task.

The whole pipeline is as under:

\begin{enumerate}
    \item We collect ~600,000 COVID-19 tweets which contain the following information: user-id, tweet, tweet-id, followers, location. 
    \item We preprocess this dataset to retain high-quality tweets. 
    \item We label the tone 12,000 tweets using IBM watson tone analyzer API. 
    \item We perform supervised fine-tuning of multiple transformer language models on this dataset (RoBERTA, BERT, XLnet, Electra).
    \item We predict the tone of 200,000 COVID19 tweets using this trained model.
    \item Lastly, we extract useful insights about the psychological condition of people throughout the timeline of COVID-19 pandemic.
    
\end{enumerate}

\begin{figure}[h!]
  \includegraphics[width=8cm]{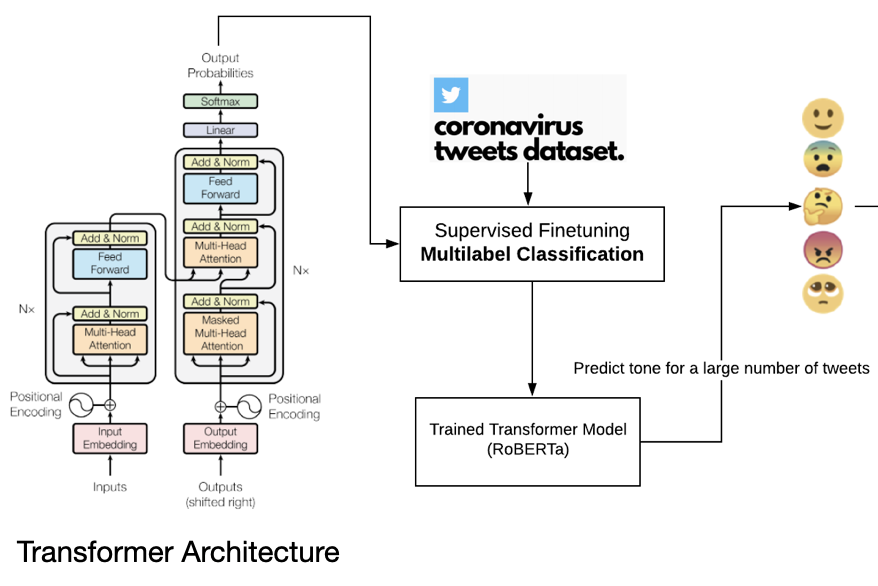}
  \caption{Methodology} 
\end{figure}

\section{Experimental Setup}

We perform supervised fine-tuning of RoBERTA, BERT-base, XLnet and Electra. Since supervised fine-tuning may take a large footprint on GPU memory, we leverage FP16 computation to reduce the size of the model. We set the learning to 0.00003. The maximum length of sequence is set to 250. The number of subbatch is set to 2. We use Adam optimizer and set the learning rate to be 3e-5. We train 3 epochs and set the gradient accumulation steps to 16. \\

Total tweets were 12,459. Train test split was 80-20.

For training the models, we use the SimpleTransformers library built on top of PyTorch. Tesla P4 GPU was used for training (on GoogleColab). 

\section{Results}
We report the Label Ranking Average Precision (LRAP) for each of the four models. \\
Given a binary indicator matrix of ground-truth labels: \\
\begin{align*}
y \in\{0,1\}^{n_{\text {samples}} * n_{\text {labels}}}
\end{align*}

The score associated with each label is denoted by $\hat{f}$ where
\begin{align*}
\hat{f} \epsilon\{\mathbb{R}\}^{n_{s a m p l e s} * n_{l a b e l s}}
\end{align*}

Then LRAP is calculated as:
\begin{align*}
L R A P(y, \hat{f})=\frac{1}{n_{\text {samples}}} * \sum_{i=0}^{n_{\text {samples}}-1} \frac{1}{\left\|y_{i}\right\|_{0}} \sum_{j: y_{i j}=1} \frac{\left|L_{i j}\right|}{\operatorname{rank} _{i j}}
\end{align*}

RoBERTa achieves the highest LRAP of 0.9267 on the test set (Table.  \ref{tab:training} ). It took 41m 8s to train, which is a reasonable training time on this dataset. The fastest transformer model was Electra which took only ~16m for supervised fine-tuning, although this came at the cost of a reduced final LRAP of 0.8553. 

\begin{figure}[h!]
  \includegraphics[width=8cm]{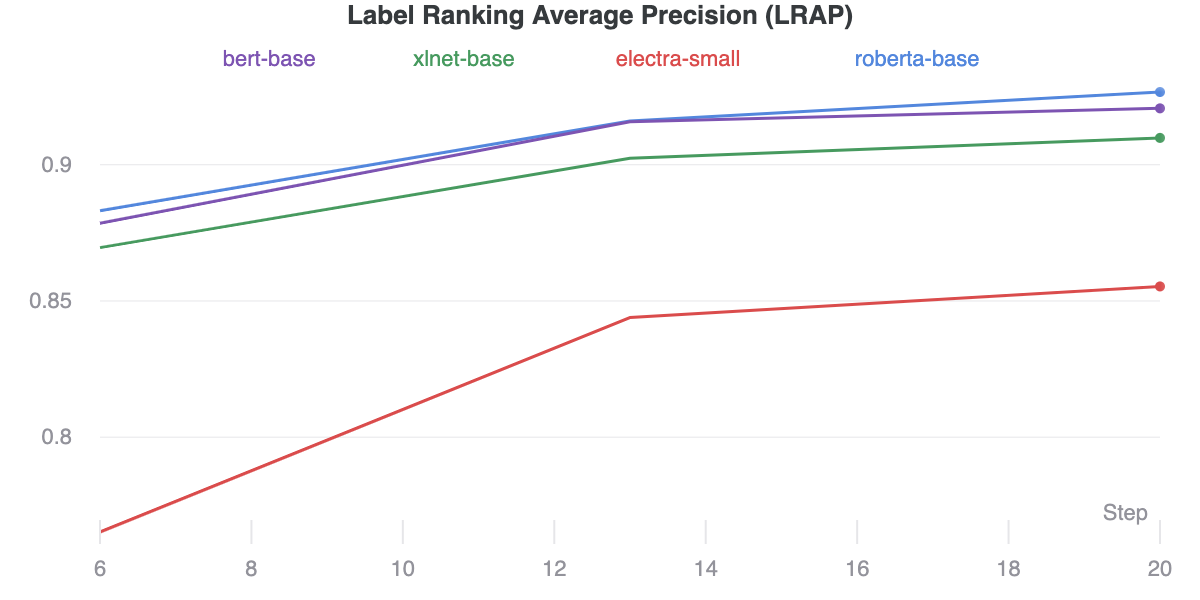}
  \caption{LRAP per step.} 
\end{figure}

\begin{figure}[h!]
  \includegraphics[width=8cm]{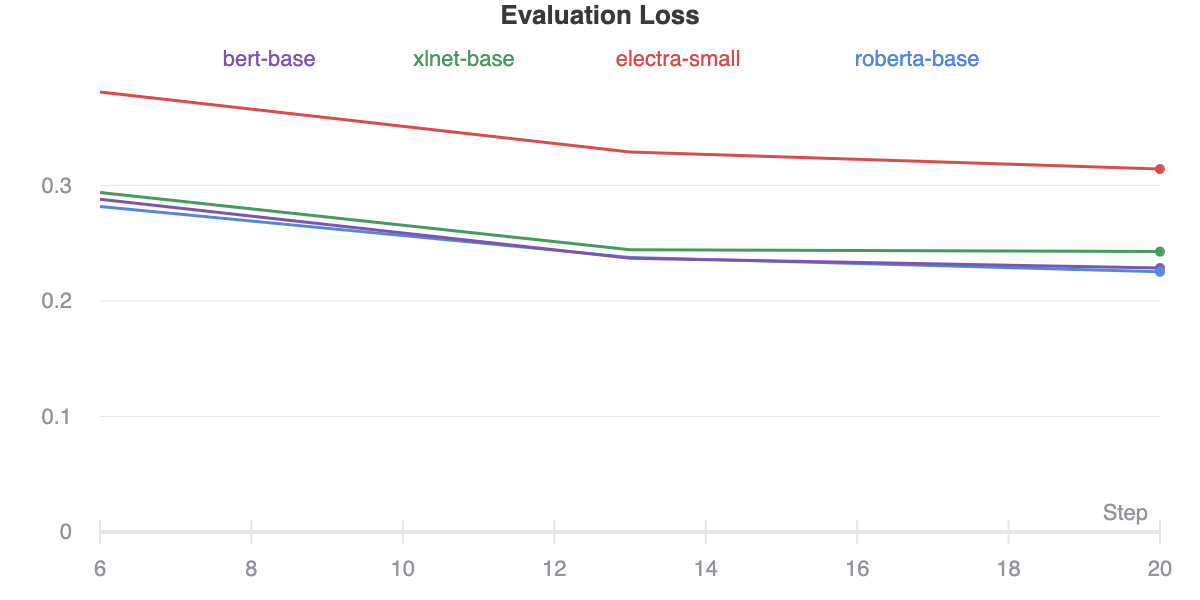}
  \caption{Loss on the evaluation set per iteration.} 
\end{figure}

\begin{table}
\begin{center}
\begin{tabular}{|l|c|c|c|}
\hline
Method & Running Time & LRAP & Eval Loss \\
\hline\hline
RoBERTa & 41m 8s & \textbf{0.9267} & \textbf{0.2254} \\
BERT & 40m 22s & 0.9207 & 0.2285 \\
XLNet & 1h 8m 40s & 0.9099 & 0.2428 \\ 
Electra Small & 16m 30s & 0.8553 & 0.3142 \\
\hline
\end{tabular}
\end{center}
\caption{Comparison of fine-tuning four transformer language models on the task of multi-label tweet classification into seven tone classes. RoBERTa achieves the highest Label Ranking Average Precision (LRAP) score of 0.9267. Electra Small was the fastest to train(16m30s) whereas XLNet took the longest time to train (1h 8m) }
\label{tab:training}
\end{table}

\section{Extracting Tone and Location}
We label the tones of 658,967 tweets using the trained RoBERTa model. The output of the model is a vector of probabilities for each of the seven tone classes. For each tweet, this vector is thresholded at 0.5, i.e. a tweet is classified as belonging to a certain tone only if that tone has been predicted with a probability greater than 0.5. This retains tones that have been predicted with high confidence. \\ \\
Next, we extract the location of users by geoparsing the location text in their profiles. We drop the tweets authored by users who have no location information. The resulting location tagged tweets are 233,762 in number. Division of these tweets amongst the seven tones is given in Table. \ref{tab:numtweets2}.

\begin{table}
\begin{center}
\begin{tabular}{|l|c|}
\hline
Tone & Number of Tweets\\
\hline\hline
Anger & 2997 \\
Analytical & 58717 \\
Tentative & 59850 \\
Confident & 34293 \\
Sadness & 23050 \\
Fear & 1163\\
Joy & 58632\\
\hline
\end{tabular}
\end{center}
\caption{Number of tweets belonging to each tone in the final labelled dataset of 233,762 tweets. }
\label{tab:numtweets2}
\end{table}

\section{Tweets Analysis}
We perform analysis on the resulting dataset containing tones and location corresponding to each tweet. \\ We define two simple mood indicators for the people of a particular country: Happiness Indicator (HI) and Sadness Indicator (SI):

\begin{align*}
HI = \frac{Number of Tweets with Joy Tone}{Total Number of Tweets}
\end{align*}

\begin{align*}
SI = \frac{Number of Tweets with Sad Tone}{Total Number of Tweets}
\end{align*}

These two indicators essentially describe the ratio of the sad or joyful tweets in a particular country.\\

\textbf{Happiest Countries}
We utilize HI to rank countries in the order of how joyful the tweets were in that country between the time-period of 25th-31st March. The results are shown in table 4. \\

\textbf{Saddest Countries}
We utilize SI to rank countries in the order of how sad the tweets were in that country between the time-period 25th-31st March. The results are shown in table 5. \\

We hypothesize from these results that the psychological condition of the people within a particular country in the early phases of COVID19 outbreak (25th to 31st March) seems to be greatly influenced by the perceived level of healthcare facilities or the ability of that country to cope with such a pandemic. The countries included in table 5 (sad sentiment) can be mostly characterized as having lesser perceived ability to deal with a healthcare crises  (Botswana, Namibia, Canada, Zimbabwe, Tonga). A study of tweets on a larger time period (e.g. Jan to July) will be be better able to confirm this hypothesis.  

\textbf{Temporal Analysis of Tweets}
We visualize the country-wise tone of tweets on a temporal chart (Figure 4.) We can see the varied sentiment of people in each of the six different countries. The most popular tones are: [Analytical, Tentative, Joy].

\begin{table}
\begin{center}
\begin{tabular}{|l|l|r|r|r|}
\hline
No.& Country & Joy &  Sadness &  HI \\
\hline\hline
1  &                          Spain &   417 &       90 &                    4.63 \\
2  &                        Germany &  1158 &      261 &                    4.43 \\
3  &                         France &   595 &      153 &                    3.88 \\
4  &                 Cayman Islands &   356 &       94 &                    3.78 \\
5  &                          Ghana &   737 &      205 &                    3.59 \\
6  &                        Ireland &  1996 &      577 &                    3.45 \\
7  &                Holy See&           713 &   210 &                       3.39 \\
8  &                  New Caledonia &  1005 &      298 &                    3.37 \\
9  &                       Mongolia &   443 &      132 &                    3.35 \\
10 &                          Macao &   351 &      105 &                    3.34 \\
\hline

\end{tabular}
\end{center}
\caption{Countries ranked according to the happiness indicator of tweets }
\label{tab:happiest}
\end{table}

\begin{table}
\begin{center}
\begin{tabular}{|l|l|r|r|r|}
\hline
No.& Country & Sadness &  Joy &  SI \\
\hline\hline
1  &     Botswana &       48 &   52 &                    0.92 \\
2  &      Namibia &       53 &   72 &                    0.73 \\
3  &        Kenya &      619 &  894 &                    0.69 \\
4  &       Zambia &       47 &   68 &                    0.69 \\
5  &      Iceland &       82 &  124 &                    0.66 \\
6  &        Japan &       75 &  115 &                    0.65 \\
7  &     Zimbabwe &       76 &  131 &                    0.58 \\
8  &        Nepal &       48 &   86 &                    0.55 \\
9  &        Tonga &      168 &  304 &                    0.55 \\
10 &       Norway &       88 &  162 &                    0.54 \\
\hline

\end{tabular}
\end{center}
\caption{Countries ranked according to the sadness indicator of tweets  }
\label{tab:saddest}
\end{table}

\begin{figure}[htp]
  \includegraphics[width=8cm]{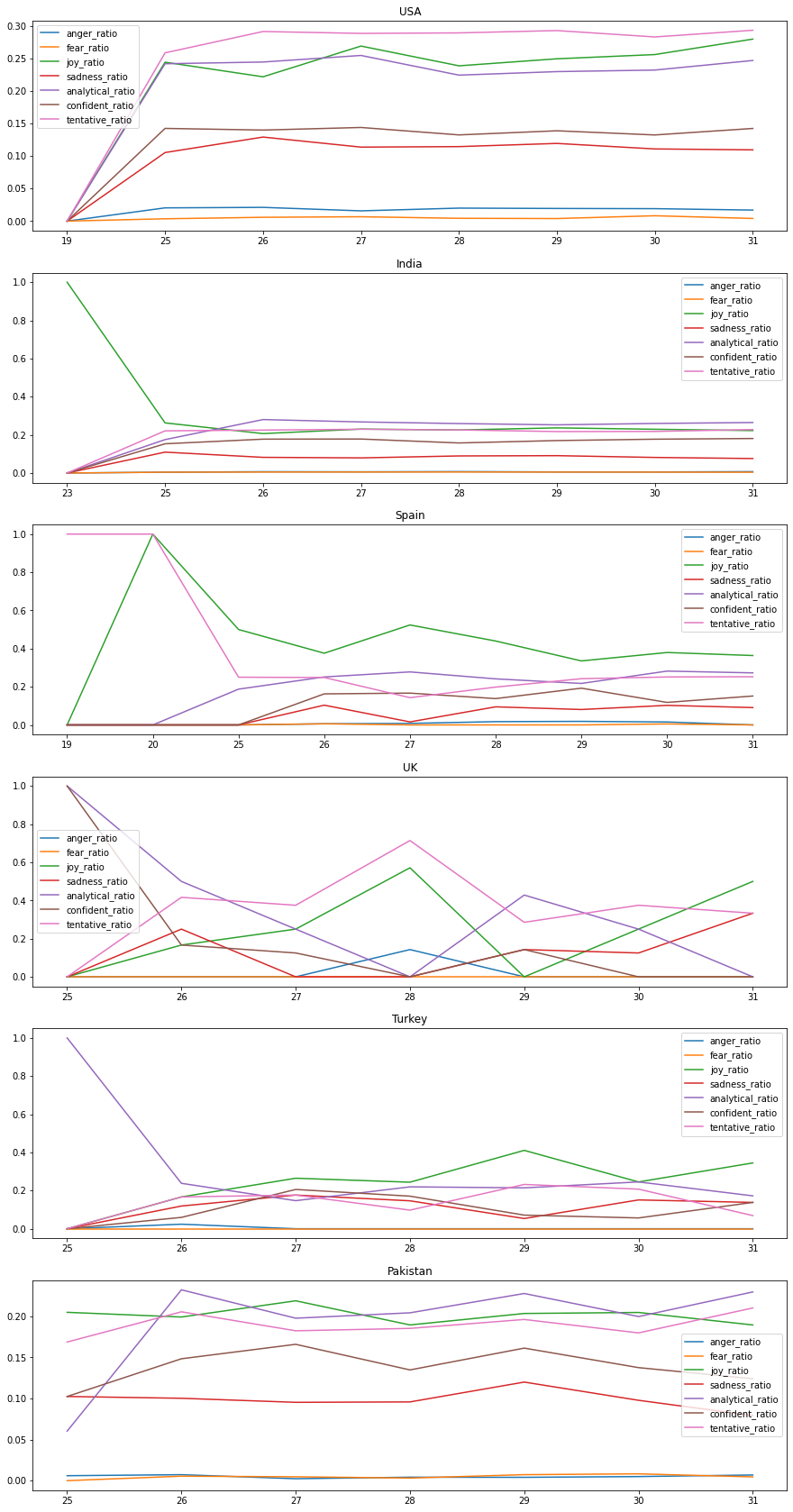}
  \caption{Temporal analysis of the tone of tweets in each of 6 countries} 
\end{figure}

\section{Conclusion}
In this project, we trained a transformer based language model, RoBERTa, on the task of multilabel tone classification in tweets. After achieving a LRAP score of 0.92, we used the trained transformer model for predicting the tone of a large number of tweets. The resulting tweets were used for performing a country-wise analysis. We hypothesized from these results that the psychological condition of the people within a particular country in the early phases of COVID19 outbreak (25th to 31st March) seems to be greatly influenced by the perceived level of healthcare facilities or the ability of that country to cope with such a pandemic. The countries ranking higher in sad sentiment can be mostly characterized as having lesser perceived ability to deal with a healthcare crises  (Botswana, Namibia, Canada, Zimbabwe, Tonga). A study of tweets on a larger time period (e.g. Jan to July) will be be better able to confirm this hypothesis.


{\small
\bibliographystyle{ieee}
\bibliography{egbib}
}

\end{document}